\documentclass[sigconf]{acmart}
\pdfoutput=1
\usepackage{hyperref}
\usepackage[normalem]{ulem}
\usepackage{graphicx}
\usepackage{caption}
\usepackage{subfig}
\usepackage{amssymb}
\usepackage{amsmath}
\usepackage{bm}
\usepackage{multirow}
\usepackage{booktabs}
\usepackage{moresize}
\usepackage{tabu}
\usepackage[lined,boxed,commentsnumbered]{algorithm2e}

\newcommand\tr{\rule{0pt}{3ex}}
\newcommand\br{\rule[-2ex]{0pt}{2ex}}
\newcommand{\ie}{\emph{i.e.}}
\newcommand{\eg}{\emph{e.g.}}
\newcommand{\etc}{\emph{etc.}}
\newcommand{\etal}{\emph{et al.}}
\newcommand{\vs}{\emph{vs.}}
\newcommand{\rf}{{\sf RF}}

\newcommand{\lr}{{\sf LogReg}}
\newcommand{\dt}{{\sf DT}}
\newcommand{\gbdt}{{\sf GBDT}}

\newcommand{\R}{\mathbb{R}}

\DeclareMathOperator*{\argmin}{arg\,min}

\setcopyright{acmcopyright}

\acmDOI{10.1145/3097983.3098039}

\acmISBN{978-1-4503-4887-4/17/08}

\acmConference{KDD '17}{August 13-17, 2017}{Halifax, NS, Canada}
\copyrightyear{2017}
\acmYear{2017}
\acmPrice{15.00}

\begin{document}
%


\title{Interpretable Predictions of Tree-based Ensembles via\\ Actionable Feature Tweaking}

\author{Gabriele Tolomei}
\affiliation{
       \institution{Yahoo Research}
       \city{London, UK}
}
\email{gtolomei@yahoo-inc.com}

\author{Fabrizio Silvestri}
\authornote{The author has contributed to this work while he was employed at Yahoo Research.}
\affiliation{
       \institution{Facebook}
       \city{London, UK}
}
\email{fsilvestr@fb.com}

\author{Andrew Haines}
\affiliation{
       \institution{Yahoo Research}
       \city{London, UK}
}
\email{haines@yahoo-inc.com}

\author{Mounia Lalmas}
\affiliation{
       \institution{Yahoo Research}
       \city{London, UK}
}
\email{mounia@acm.org}

\begin{abstract}

Machine-learned models are often described as ``black boxes''.
In many real-world applications however, models may have to sacrifice predictive power in favour of human-interpretability.
When this is the case, feature engineering becomes a crucial task, which requires significant and time-consuming human effort. Whilst some features are inherently static, representing properties that cannot be influenced (\eg, the age of an individual), others capture characteristics that could be \emph{adjusted} (\eg, the daily amount of carbohydrates taken). 
Nonetheless, once a model is learned from the data, each prediction it makes on new instances is irreversible - assuming every instance to be a static point located in the chosen feature space.
There are many circumstances however where it is important to understand \emph{(i)} why a model outputs a certain prediction on a given instance, \emph{(ii)} which adjustable features of that instance should be modified, and finally \emph{(iii)} how to alter such a prediction when the mutated instance is input back to the model.

In this paper, we present a technique that exploits the internals of a tree-based ensemble classifier to offer \emph{recommendations} for transforming true negative instances into positively predicted ones. 
We demonstrate the validity of our approach using an online advertising application. 
First, we design a Random Forest classifier that effectively separates between two types of ads: \emph{low} (negative) 
and \emph{high} (positive) quality ads (instances). 
Then, we introduce an algorithm that provides recommendations that aim to transform a low quality ad (negative instance) into a high quality one (positive instance). 
Finally, we evaluate our approach on a subset of the active inventory of a large ad network, \emph{Yahoo Gemini}.

\end{abstract}
%
%



\maketitle

\section{Introduction}
\label{sec:intro}

An increasing number of organisations and governments rely on \emph{Machine Learning} (ML) techniques to extract knowledge from the large volumes of data they collect every day to optimise their operational effectiveness.

ML solutions are usually considered as ``black boxes''; they take some inputs and produce desired outputs, especially when their ultimate goal is \emph{prediction} rather than \emph{inference}. As long as ML models work properly, ``everybody'' is happy and little attention is devoted to understand why such surprisingly good results are obtained. 
Still, it is beneficial to have available techniques supporting humans in interpreting and ``debugging'' these models, particularly when they fail~\cite{szegedy2013intriguing} or lead to some oddities.\footnote{\url{http://www.telegraph.co.uk/technology/2016/03/24/microsofts-teen-girl-ai-turns-into-a-hitler-loving-sex-robot-wit/}}

Excluding recent trends in ML such as Deep Learning~\cite{lecun2015nature}, typically the initial effort when designing an ML solution consists in modelling the objects of a given domain of interest, \ie, \emph{feature engineering}.
This step aims to describe each object in the domain using an appropriate set of properties (\emph{features}), which define a so-called \emph{feature space}.
For a given dataset, each object can be considered as a static point located in the feature space since each feature value is deemed to be fixed; once a model is learned from the data, each prediction it makes on new objects is irreversible.

Let us assume that we disagree with a prediction that the model returns for a given object or that we would like to enforce switching such a prediction. The research question we ask in this work is \emph{how can we understand what can be changed in the feature vector in order to modify the prediction accordingly?}

To better understand this challenge with an example, consider an ML application in the healthcare domain, where patients (objects) are mapped to a vector of clinical indicators (features), such as age, blood pressure, daily carbohydrates taken, \etc\ Assume next that an ML model has been designed to accurately predict from these features whether a patient is at risk of a heart attack or not.
If for a given patient our model predicts that there is a high risk of a heart attack it would be of great advantage for medical physicians to also have a tool that suggests the most appropriate clinical treatment by offering targeted adjustments to specific indicators (\eg, reducing the daily amount of carbohydrates). In other words, to recommend the clinical treatment to switch a patient from being of high risk (negative instance) to low risk (positive instance).

In this work, we propose an algorithm for \emph{tweaking} input features so as to change the output predicted by an existing machine-learned model.
Our method is designed to operate on top of any tree-based ensemble binary classifier, although it can be extended to multi-class classification. Our proposed algorithm exploits the internals of the model to generate recommendations for transforming true negative instances into positively predicted ones (or vice versa).

We describe the theoretical framework along with experiments designed to validate the proposed algorithm. 
Our approach is then evaluated in the commercial and more implementable setting of online advertisement recommendations to illustrate the generic nature of our framework and the many and varied domains it can be applied to.

After presenting an effective Random Forest classifier that is able to separate between \emph{low} and \emph{high} quality advertisements~\cite{Lalmas:2015:PPP:2783258.2788581}, we show how our algorithm can be used to automatically generate ``interpretable'' and ``actionable'' suggestions on how to convert a low quality ad (negative instance) into a high quality one (positive instance).
Such insights can be provided to advertisers who may turn them into actual changes to their ad campaigns with the aim of improving their return on investment.
Finally, we assess the quality of recommendations that our algorithm generates out of a dataset of advertisements served by the \emph{Yahoo Gemini} ad network. 

\section{Problem Statement}
\label{sec:problem}  

We start by considering the typical binary classification problem and focus on an ensemble of tree-based classifiers as an effective solution to the above problem.
Additionally, we define how the internals of an existing ensemble of trees can be used to derive a feedback loop for recommending how true negative instances can be turned into positively predicted ones (or vice versa). 

The approach we propose can be easily extended to the more general multi-class classification problem. We plan to present this result in future extended work.

\subsection{Notation}
\label{subsec:notation}
Let $\mathcal{X}\subseteq \R^n$ be an $n$-di\-men\-sio\-nal vector space of real-valued features.
Any $\mathbf{x}\in \mathcal{X}$ is an $n$-di\-men\-sio\-nal feature vector, \ie, $\mathbf{x} = (x_1, x_2, \ldots, x_n)^T$, representing an object in the vector space $\mathcal{X}$.
Suppose that each $\mathbf{x}$ is associated with a binary \emph{class label} -- either {\tt neg} ({\em negative}) or {\tt pos} ({\em positive}) -- and let $\mathcal{Y} = \{-1,+1\}$ be the set encoding all such possible class labels.

We assume there exists an \emph{unknown target} function $f: \mathcal{X} \longmapsto \mathcal{Y}$ that maps any feature vector to its corresponding class label.
In addition, we let $\hat{f}\approx f$ which is learned from a labelled dataset of $m$ instances $\mathcal{D} = \{(\mathbf{x_1}, y_1), (\mathbf{x_2}, y_2), \ldots, (\mathbf{x_m}, y_m)\}$.
More specifically, $\hat{f}$ is the estimate that best approximates $f$ on $\mathcal{D}$, according to a specific \emph{loss function} $\ell$.
Such a function measures the ``cost'' of prediction errors we would make if we replaced the true target $f$ with the estimate $\hat{f}$.

The flexibility \vs\ interpretability of $\hat{f}$ depends on the \emph{hypothesis space} which $\hat{f}$ has been picked from by the learning algorithm.
In this work, we focus on $\hat{f}$ represented as an \emph{ensemble} of $K$ tree-based classifiers, $\hat{f} = \phi(\hat{h}_1,\ldots, \hat{h}_K)$. Each $\hat{h}_k:\mathcal{X} \longmapsto \mathcal{Y}$ is a base estimate, and $\phi$ is the function responsible for combining the output of all the individual base classifiers into a single prediction.

A possible implementation of $\phi$ could use a \emph{majority voting} strategy. 
In this setting, a given instance $\mathbf{x}$ would obtain a predicted class label $\hat{f}(\mathbf{x})$ based on the result of the majority of the base classifiers; this is the \emph{mode} of the base predictions.
Although other strategies may be used, this does not impact our proposed approach.

\subsection{Enforcing Positive Prediction}
\label{subsec:enforce-prediction}

In any ensemble of tree-based classifiers, each base estimate $\hat{h}_k$ is encoded by a decision tree $T_k$, and the ensemble is represented as a forest $\mathcal{T} = \{T_1, \ldots, T_K\}$.

Our aim is to identify how to transform a true negative instance into a positively predicted one. 
Let $\mathbf{x}\in \mathcal{X}$ be a true negative instance such that $f(\mathbf{x}) = \hat{f}(\mathbf{x}) = -1$.
The task can now be defined as transforming the original input feature vector $\mathbf{x}$ into a new feature vector $\mathbf{x'}$ ($\mathbf{x} \leadsto \mathbf{x'}$) such that $\hat{f}(\mathbf{x'}) = +1$. 
Moreover, we accomplish an optimised form of the problem by choosing $\mathbf{x'}$ as the best transformation among all the possible transformations $\mathbf{x^*}$, according to a \emph{cost function} $\delta: \mathcal{X}\times \mathcal{X} \longmapsto \R$.
This is defined as follows:
\[
\mathbf{x'} = \argmin_{\mathbf{x^*}} \Big\{\delta(\mathbf{x}, \mathbf{x^*})~|~\hat{f}(\mathbf{x}) = -1 \wedge \hat{f}(\mathbf{x^*}) = +1 \Big\}
\]
The cost function measures the ``effort'' of transforming $\mathbf{x}$ into $\mathbf{x'}$.
A possible choice of such a function is the number of features affected by the transformation or the Euclidean distance between the original and the transformed vector.  

\subsection{Positive and Negative Paths}
\label{subsec:posneg-paths}
Any root-to-leaf path of a single decision tree can be interpreted as a cascade of \emph{if}-\emph{then}-\emph{else} statements, where every internal (non-leaf) node is a boolean test on a specific feature value against a threshold. We restrict the tree decisions to be binary representations as any multiway decision can be represented in a binary form and there is little performance benefit in n-ary splits. 
An instance's feature value is then evaluated at each node to determine which branch to traverse. This is repeated until the leaves are reached whereby the {\tt pos}/{\tt neg} classification labels are defined and assigned.

Given a forest of $K$ decision trees $\mathcal{T} = \{T_1, \ldots, T_K\}$, we denote by $p_{k,j}$ the $j$-th path of the $k$-th tree $T_k$.
We refer to $p^{+}_{k,j}$ (or $p^{-}_{k,j}$) as the $j$-th path of $T_k$ that leads to a leaf node labelled as {\tt pos} (or {\tt neg}) -- a \emph{positive} (or \emph{negative}) path.
For simplicity, we assume that each path of a decision tree contains at most $n$ non-leaf nodes, which correspond to $n$ boolean conditions, one for each distinct feature.\footnote{\small{In general, there can be multiple boolean conditions associated with a single feature.}} We thus represent a root-to-leaf path as follows:
\begin{equation}
	\label{eq:path}
	p_{k,j} = \{(x_1 \lesseqgtr \theta_1), (x_2 \lesseqgtr \theta_2), \ldots, (x_n \lesseqgtr \theta_n)\}
\end{equation}  
Let $P^{+}_k = \bigcup_{j\in T_k}p^{+}_{k,j}$ describe the set of all positive paths, and $P^{-}_k = \bigcup_{j\in T_k}p^{-}_{k,j}$ the set of all the negative paths in $T_k$. Also, let $P_k = P^{+}_k \cup P^{-}_k$ be the set of \emph{all} the paths in $T_k$.

We thus enumerate the possible paths in a single decision tree.
Even under the assumption that each $p_{k,j}\in P_k$ is at most a length-$n$ path then $T_k$ is a depth-$n$ binary tree, whose number of leaves is therefore bounded to~$2^n$. 
As the total number of leaves coincides with the total number of possible paths, we obtain $|P_k| \leq 2^n$.

In general, we cannot ensure a bound on each $|P_k|$, as there might exist some paths $p_{k,j}\in P_k$ whose length is greater than $n$.
In practice though, we can specify the maximum number of paths at training time by bounding the depth of the generated trees to the number $n$ of features.
Even with such a relaxed condition, the total number of possible paths encoded by the forest $\mathcal{T}$ is equal to $\sum_{k=1}^K |P_k| \leq K2^n$, therefore still exponential in $n$.
We see later how this does not disrupt computational efficacy in practice, as our algorithm operates on a subset of those paths.

\subsection{Tweaking Input Features}
\label{subsec:tweaking}
Given our input feature vector $\mathbf{x}$, we know from our hypothesis that $f\left(\mathbf{x}\right) = \hat{f}(\mathbf{x}) = -1$.
If the overall prediction is obtained using a majority voting strategy, it follows:
\[
\hat{f}(\mathbf{x}) = -1 \iff \left( \sum_{k=1}^K \hat{h}_k(\mathbf{x})\right) \leq 0
\]
Furthermore, there must be at least $\left\lceil\frac{K}{2}\right\rceil$ decision trees (base classifiers) of the forest $\mathcal{T}$ whose output is $-1$. That is, there exists $K^- \subseteq \{1,\ldots,K\}$ with $|K^-| \geq \left\lceil\frac{K}{2}\right\rceil$, such that:
\[
\hat{h}_{k^-}(\mathbf{x}) = -1,~\forall k^-\in K^-
\]
As we are operating in a binary classification setting, there must also exist $K^+ = \{1,\ldots,K\}\setminus K^-$, which denote the set of classifier indices that output a positive label when input with $\mathbf{x}$, \ie, $\hat{h}_{k^+}(\mathbf{x}) = +1,~\forall k^+\in K^+$.

Our goal is to tweak the original input feature vector $\mathbf{x}$ so as to adjust the prediction made by the ensemble from negative ($-1$) to positive ($+1$).
We can skip all the trees indexed by $K^+$, as these are already encoding the (positive) prediction we ultimately want. 
We therefore focus on each tree $T_k$ where $k \in K^-$, and consider the set $P^{+}_k$ of all its positive paths.
With each $p^{+}_{k,j}\in P^{+}_k$ we associate an instance $\mathbf{x}^+_{j} \in \mathcal{X}$ that \emph{satisfies} that path -- \ie, an instance whose adjusted feature values meet the boolean conditions encoded in $p^{+}_{k,j}$ to finish on a 
{\tt pos}-labelled leaf, and therefore $\hat{h}_k(\mathbf{x}^+_{j}) = +1$.

Among all the possibly infinite instances satisfying $p^{+}_{k,j}$, we restrict to $\mathbf{x}_{j(\epsilon)}^+$ to be feature value changes with a ``tolerance'' of at most $\epsilon$. We call it the $\epsilon$-satisfactory instance of $p^{+}_{k,j}$.
We consider $p^{+}_{k,j}$ containing at most $n$ boolean conditions, as specified by Equation~\ref{eq:path}.
Therefore, for any (small) fixed $\epsilon > 0$, we build a feature vector $\mathbf{x}_{j(\epsilon)}^+$ as follows:
\begin{equation}
	\label{eq:tweak}
\mathbf{x}_{j(\epsilon)}^+[i] = \left\{
    \begin{array}{l l}
      \theta_i - \epsilon & \quad \text{if the $i$-th condition is $(x_i \leq \theta_i)$}\\
      \theta_i + \epsilon & \quad \text{if the $i$-th condition is $(x_i > \theta_i)$}
    \end{array} \right.
\end{equation}
Having a single \emph{global} tolerance $\epsilon$ for all the features works as long as we standardise every feature (\eg, using z-score or min-max). If we use standard z-score for each feature, the actual magnitude of the change cannot just depend on $\epsilon$ but instead needs to be considered a multiple of a unit of standard deviation from the feature mean. Let $\theta_i = \frac{t_i - \mu_i}{\sigma_i}$ be the z-score of the threshold on the $i$-th feature value, where $t_i$ is the non-standardised value, $\mu_i$ and $\sigma_i$ are the mean and standard deviation of the $i$-th feature 
respectively.\footnote{\small{In practice, $\mu_i$ and $\sigma_i$ are often unknown, so the sample mean and the sample standard deviation are used instead.}}
Now, suppose that $x_i = \theta_i \pm \epsilon$ is the tweaked value of the $i$-th feature according to the ongoing transformation of the input vector $\mathbf{x}$.
Therefore, $x_i = \frac{t_i - \mu_i}{\sigma_i}\pm \epsilon$.
Returning to the original (\ie, non-standardised) feature scale, we obtain $x_i = t_i - \mu_i \pm \epsilon \sigma_i$.
Depending on the sign, the tweaked feature is either moving closer to or farther away from the original feature mean $\mu_i$, pivoting around $t_i$.

For each $p^{+}_{k,j} \in P^{+}_k$ we transform our input feature vector $\mathbf{x}$ into the $\epsilon$-satisfactory instance $\mathbf{x}_{j(\epsilon)}^+$ that validates $p^{+}_{k,j}$. 
This leads us to a set of transformations $\Gamma_k = \bigcup_{j\in P^{+}_k}\mathbf{x}_{j(\epsilon)}^+$, associated with the $k$-th tree $T_k$.

Each resulting transformation in $\Gamma_k$ may have an impact on other trees of the forest.
There might exist $l \in K^+$ whose corresponding tree already provides the correct prediction when this is input with $\mathbf{x}$, $\hat{h}_{l}(\mathbf{x}) = +1$.
It may also happen that by changing $\mathbf{x}$ into $\mathbf{x}'$ the prediction of the $l$-th tree is incorrect and now $\hat{h}_{l}(\mathbf{x}') = -1$.
In other words, by changing $\mathbf{x}$ into another instance $\mathbf{x}' \in \Gamma_k$ we are only guaranteed that the prediction of the $k$-th base classifier is correctly fixed, \ie, from $\hat{h}_k(\mathbf{x}) = -1$ to $\hat{h}_k(\mathbf{x}') = +1$.
The overall prediction for $\mathbf{x}'$ may or may not be fixed, where $\hat{f}(\mathbf{x}')$ may still output $-1$, exactly as $\hat{f}(\mathbf{x})$ did.

If the change from $\mathbf{x}$ to $\mathbf{x}'$ also leads to $\hat{f}(\mathbf{x}') = +1$, then $\mathbf{x}'$ will be a \emph{candidate} transformation for $\mathbf{x}$.
More formally, let $\Gamma = \bigcup_{k=1}^K \Gamma_k$ be the set of \emph{all} the $\epsilon$-satisfactory transformations of the original $\mathbf{x}$ from the positive paths of all the trees in the forest. Our feature 
tweaking problem can then be generally defined as follows:
\[
\mathbf{x'} = \argmin_{\mathbf{x}_{j(\epsilon)}^+\in \Gamma~|~\hat{f}(\mathbf{x}_{j(\epsilon)}^+) = +1} \Big\{\delta(\mathbf{x}, \mathbf{x}_{j(\epsilon)}^+) \Big\}
\]
In~\cite{cui2015kdd}, it has already been proven that a problem similar to the one we define above is NP-hard as it reduces to DNF-MAXSAT. Our version, in fact, introduces an additional constraint ($\epsilon$) on the possible way features can be tweaked and thus it is itself NP-hard.
This problem definition is still valid for the base case when $K=1$.
There, the additional condition requiring $\hat{f}(\mathbf{x}_{j(\epsilon)}^+) = +1$ is not necessary because it is implicitly true by definition. 
In that scenario, the ensemble is composed of a single base classifier -- \ie, the forest contains a single decision tree and tweaking its prediction also results in changing the overall prediction.
Note that when there is only one decision tree, our problem can be solved \emph{optimally}: We can enumerate all the positive paths, choose the one with the minimum cost, and check if the threshold of tolerance $\epsilon$ is satisfied on each feature. 
Because base trees are interconnected through the features they share, simply enumerating positive paths does not work for an ensemble of trees since the output of a base tree may affect outputs of its sibling trees.
The presence of the condition $\hat{f}(\mathbf{x}_{j(\epsilon)}^+) = +1$ circumvents this issue by querying the model itself on the correctness of the $\epsilon$-transformation of an instance. 

\subsection{The Feature Tweaking Algorithm}
\label{subsec:algorithm}

Our approach takes as input 4 key components: \emph{(i)} The trained ensemble model $\hat{f}$; \emph{(ii)} A feature vector $\mathbf{x}$ that represents a true negative instance; \emph{(iii)} A cost function $\delta$ measuring the ``effort'' required to transform the true negative instance into a positive one; and 
\emph{(iv)} A positive threshold $\epsilon$ that bounds the tweaking of each single feature to pass every boolean test on a positive path of each tree. 
The result being the transformation $\mathbf{x'}$ of the original $\mathbf{x}$ that exhibits the minimum cost according to $\delta$.
The detailed description is presented in Algorithm~\ref{algo:tweakfeat}.

\begin{algorithm}[t]
\small
 	\SetAlgoLined
	\SetKwFunction{getPositivePaths}{getPositivePaths}
	\SetKwFunction{buildPositiveInstance}{buildPositiveInst}
	\SetKwInOut{Input}{Input}
	\SetKwInOut{Output}{Output}
	\SetCommentSty{emph}
	
	\Input{\\
	\noindent $\triangleright$ An estimate function $\hat{f}$ resulting from an ensemble of decision trees $\mathcal{T} = \{T_1,\ldots,T_K\}$, each one associated with a base estimate $\hat{h}_k,~k=1,\ldots,K$\\
	\noindent $\triangleright$ A feature vector $\mathbf{x}$ representing a \emph{true negative} instance, such that $f(\mathbf{x}) = \hat{f}(\mathbf{x}) = -1$\\
	\noindent $\triangleright$ A cost function $\delta$\\
	\noindent $\triangleright$ A (small) threshold $\epsilon > 0$\\
	}
 	\Output{\\
	\noindent $\triangleright$ The optimal transformation $\mathbf{x}'$ with respect to $\delta$, such that $\hat{f}(\mathbf{x}') = +1$
	}

 	\BlankLine
 	\Begin{
  		$\mathbf{x}' \longleftarrow \mathbf{x}$\;
		$\delta_{\textit{min}} \longleftarrow +\infty$\;
		\For {$k=1,\ldots, K$} {
			\If {$\hat{f}(\mathbf{x}) == \hat{h}_k(\mathbf{x})~{\bf and}~\hat{h}_k(\mathbf{x}) == -1$ }{
				\tcc{retrieve the set of \emph{positive} paths of the $k$-th decision tree}
				$P^{+}_k \longleftarrow$ \getPositivePaths{$T_k$}\;
				\ForEach {$p^{+}_{k,j} \in P^{+}_k$} {
					\tcc{generate the $\epsilon$-satisfactory instance associated with the $j$-th positive path of the $k$-th decision tree}
					$\mathbf{x}_{j(\epsilon)}^+ \longleftarrow $ \buildPositiveInstance{$\mathbf{x}, p^{+}_{k,j}, \epsilon$}\;
					\If {$\hat{f}(\mathbf{x}_{j(\epsilon)}^+) == + 1$ }{
							\If {$\delta(\mathbf{x}, \mathbf{x}_{j(\epsilon)}^+) < \delta_{\textit{min}}$ }{
								$\mathbf{x}' \longleftarrow \mathbf{x}_{j(\epsilon)}^+$\;
								$\delta_{\textit{min}} \longleftarrow \delta(\mathbf{x}, \mathbf{x}_{j(\epsilon)}^+)$\;
							}
 						}
					}
				}
			}
 		\Return{$\mathbf{x}'$}\;
 	}
 	\caption{The {\sc Feature Tweaking Algorithm}}
 	\label{algo:tweakfeat}
\end{algorithm}

In the worst case, our algorithm examines all $K$ trees of the forest, although it investigates only those trees whose base predictions are negatives ({\tt neg}-labelled).
Then, all the positive paths of each tree are considered, and for each of those a potential candidate $\epsilon$-transformation is built according to the scheme proposed in Equation~\ref{eq:tweak}.
As such, the number of steps depends on the number of positive paths on each tree, which in turn is related to the number of leaves. 

We stated in Section~\ref{subsec:posneg-paths} that we cannot provide \emph{apriori} any limit to the depth of a decision tree, and therefore to its number of leaves.
However, in practice these can be bounded whilst training the model.
We therefore set the maximum length of each root-to-leaf path, \ie, the depth of each tree, to be at most equal to the number of input features $n$.
Considering our goal, this is not a limitation as each transformation should at most affect each feature exactly once.
The total number of positive paths to be examined will then be limited by $K2^n$, and so the worst case complexity is $O(2^n)$.
It follows that our method might be unsuitable for dealing with high-dimensional datasets, such as text, images, or videos.
In reality such a setting would not really make  sense when transforming input instances in the original feature space (\eg, change a few words on a document or a few pixels of an image).

Although the search space can be exponential in the number of features, in practice the number of positive paths of each tree is significantly smaller.
This makes our method feasible on average input sizes (\ie, below 100 features).
For example, in our experiments we found that the maximum depth of each tree leading to the best classification results is significantly smaller than the total number $n=45$ of features (see Table~\ref{tab:cv-performance}).
In addition, many positive paths may share several boolean conditions, especially when extracted from the same decision tree.
This allows us to avoid tweaking the same input feature multiple times according to the same condition by using some caching mechanism.

Finally, our algorithm can be easily parallelised since each tree can be explored independently from the others for any given instance --\ie, we can adjust the $k$-th tree while keeping the remaining $K\setminus k$ trees simultaneously fixed for all $k\in \{1,\ldots, K\}$.


\section{Use Case: Improving Ad Quality}
\label{sec:adquality-prediction}

We demonstrate the utility of our method when applied to a real-world use case in online advertising.
We investigate how our algorithm can be used to improve the \emph{quality} of advertisements served by the \emph{Yahoo Gemini} ad network.

\subsection{Why Ad Quality?}
\label{subsec:adquality-why}
A main source of monetisation for online web services comes in the form of advertisements (\emph{ads} for short)  impressed in dedicated real estate units of rendered web pages. 
Online publishers operating these web services typically reserve predefined slots within their streams, utilising a third party ad network to deliver ad inventory to impress within them. 
Ad networks free publishers from running their own ad servers as they decide for them which ads should be placed at which slots, when and to whom.
Advertisers rely on ad networks to optimise their \emph{return on investment} -- for example through targeting the right audience according to the advertiser budget and marketing strategy.

A trustworthy relationship between advertisers, publishers and ad networks is instrumental to the success of online advertising.
On the advertiser side, ad networks provide advertisers with tools for monitoring key performance indicators (\eg, number of ad \emph{impressions}, \emph{click-through rate} or CTR, \emph{bounce rate}) as well as effective mechanisms to overcome fraudulent activities such as \emph{click spam}~\cite{Daswani2007AC, StoneGross2011UFA}. 
On the publisher side, ad networks provide mechanisms that allow users to hide ads they dislike and indicate the reasons for doing so. 
This information can be used by ad networks to ensure that the ads they serve do not negatively affect user engagement on a publisher website, as well as reporting to advertisers on the \emph{quality} of their ads.

As with any self served content delivery platform, an ad network has a varying distribution of quality with many ads being of low quality. Not serving them may not be an option when supply is exhausted. 
Therefore, another approach to positively shift the inventory quality distribution is to leverage the interpretability of the internal machinery of existing ad quality prediction models -- \ie, binary classifiers -- so as to offer actionable \emph{recommendations}. The intention of these programmatically-computed recommendations is to provide advertisers with guidance on how they can improve their ad quality at scale. Such a system yields value for all beneficiaries in this advertising ecosystem ultimately culminating with a better user experience.

Returning to our work, by applying our feature tweaking algorithm introduced in Section~\ref{subsec:algorithm}, we show in the rest of this paper how we can transform a low quality ad into a set of new ``proposed'' high quality ads by shifting their position in an ad quality feature space. 
The algorithm employs the internals of a learned binary classifier to \emph{tweak} the feature-based representation of a low quality ad so that the new ``proposed'' ads are promoted to high quality ones when 
re-input to the classifier. 
Each transformation is associated with a \emph{cost}, allowing us to generate actionable suggestions from the ``proposed'' instances with the least cost, so as to improve low quality ads.
We validate our approach on a dataset of mobile native ads served by the \emph{Yahoo Gemini}\footnote{\small{\url{https://gemini.yahoo.com}}} ad network.

\subsection{A Definition of Ad Quality}
\label{subsec:adquality-definition}
Many factors can affect the quality of an ad: its \emph{relevance}, \ie, whether the ad matches the user interest~\cite{Raghavan:2009:RMB:1571941.1572116}; the \emph{pre-click} experience, \ie, whether the ad annoys a user~\cite{zhou2016predicting}; 
and finally the \emph{post-click} experience, \ie, whether the \emph{ad landing page}\footnote{\small{We refer to \emph{ad landing page} as the web page of the advertiser that a user is redirected to after clicking on an \emph{ad}.}} 
meets the user click intent that brought them to the landing page~\cite{Lalmas:2015:PPP:2783258.2788581}. 
We focus on the latter, the post-click experience, following from ~\cite{DBLP:conf/www/BarbieriSL16,Lalmas:2015:PPP:2783258.2788581}.

Inspired by these studies, we define and measure the quality of ads using the time spent on their landing pages as a proxy, referred to as \emph{dwell time}. 
We know from~\cite{Lalmas:2015:PPP:2783258.2788581} that ad landing pages exhibiting long dwell times promote a positive long-term post-click experience. Based on this definition of ad quality, we design a binary classifier that effectively 
separates between \emph{low} and \emph{high} quality ads, \ie, ad landing pages whose dwell time is below or above a threshold $\tau$, respectively.
We compute $\tau$ as the median of all the sample means of dwell times observed for a large set of ad landing pages.
Intuitively, an ad landing page is of high quality if its average dwell time is greater than $\tau$ --
\ie, if the average time users spent on the page is greater than the average time users spent on at least 50\% of any other ad landing page.
Although more sophisticated approaches can be designed~\cite{DBLP:conf/www/BarbieriSL16}, this is not the main goal of this research, and we leave it for future work.

\subsection{Predicting Ad Quality}
\label{subsec: adquality-prediction}
To apply our algorithm to our use case, we first need to learn a binary classifier that predicts whether an ad is of high quality or not, given a feature-based representation of the ad creative and the landing page.

\subsubsection{Ad Feature Engineering}
\label{subsubsec:features}

A sample of the set of features used in this work is listed in Table~\ref{tab:features}.
This is based on the same set used in~\cite{DBLP:conf/www/BarbieriSL16}, with the additional ``{\sf Language}'' category (marked with ``$\dagger$'' in the Table).
Due to space constraints we do not show all the features, and invite the reader to refer to~\cite{DBLP:conf/www/BarbieriSL16} for a complete  description of them. 

Each feature in the table is associated with a \emph{category} and a \emph{source}.
The former indicates the type of features whilst the latter specifies whether the feature is computed from the ad \emph{landing page} ({\sf LP}), the ad \emph{creative} ({\sf CR}), or a combination of the two ({\sf CR-LP}).
Although our focus is on the post-click experience (\ie, ad landing page), we aim to obtain the best performing model for predicting the quality of the ads, and hence include both pre-click (\ie, ad creative) and historical features; the latter not participating in the tweaking process as -- by definition -- they cannot be altered.

\begin{table*}[!htb]
\caption{\label{tab:features} {\small Set of features used to characterise ad \emph{creative} ({\sf CR}), ad \emph{landing page} ({\sf LP}), or both ({\sf CR-LP}).}}

\small
\begin{center}
\begin{tabu} to \textwidth {| X[2,c,m] | X[1,c,m] |  X[5,c,m] |}
	\hline
	\tr{\bf Category}\br & \tr{\bf Source}\br & \tr{\bf Description}\br
 	\\ \hline
 	\multirow{1}{*} {\sf Language}{$^\dagger$} & {\sf CR} & {This set of features capture the extent to which the text of the ad creative may include adult, violent, or spam content (\eg, ADULT\_SCORE, HATE\_SCORE, and SPAM\_SCORE)}
 	\\ \hline
    \multirow{1}{*} {\sf DOM} & {\sf LP} & {This set of features are derived from the elements extracted from the HTML DOM of the ad landing page, such as the main textual content (LANDING\_MAIN\_TEXT\_LENGTH), the total number of internal and external hyperlinks (LINKS\_TOTAL\_COUNT), the ratio of main text length to the total number of hyperlinks on the page (LINKS\_MAIN\_LENGTH\_TOTAL\_RATIO), \etc}
    \\ \hline
    \multirow{1}{*} {\sf Readability} & {\sf CR-LP} & {These features range from a simple count of tokens (words) in the text of the ad creative and landing page to well-known scores for measuring the summarisability/readability of a text (\eg, READABILITY\_SUMMARY\_SCORE), \etc}
\\ \hline
    \multirow{1}{*} {\sf Mobile Optimising} & {\sf LP} & {This set of features describe the degree of mobile optimisation of the ad landing page by measuring the ability of it to be tuned to different screen sizes (VIEW\_PORT), testing for the presence of a click-to-call button (CLICK\_TO\_CALL), \etc}
\\ \hline
    \multirow{1}{*} {\sf Media} & {\sf LP} & {These features refer to any media content displayed within the ad landing page, such as the number of images (NUM\_IMAGES), \etc}
\\ \hline
    \multirow{1}{*} {\sf Input} & {\sf LP} & {This set of features represent all the possible input types available on the ad landing page, such as the number of checkboxes, drop-down menus, and radio buttons (NUM\_INPUT\_CHECKBOX, NUM\_INPUT\_DROPDOWN, NUM\_INPUT\_RADIO), \etc}
\\ \hline
    \multirow{1}{*} {\sf Content \& Similarity} & {\sf CR-LP} & {These features extract the set of Wikipedia \emph{entities} from the ad creative and landing page (NUM\_CONCEPT\_ANNOTATION), and measure the \emph{Jaccard} similarity between those two sets (SIMILARITY\_WIKI\_IDS), \etc}
\\ \hline
    \multirow{1}{*} {\sf History} & {\sf LP} & {These features measure historical indicators, such as the median \emph{dwell time} as computed from the last 28 days of observed ad clicks (HISTORICAL\_DWELLTIME), and the \emph{bounce rate} -- \ie, the proportion of ad clicks whose dwell time is below 5 seconds (HISTORICAL\_BOUNCE\_RATE), \etc }
	\\ \hline
	\end{tabu}
\end{center}
\end{table*}

\subsubsection{Learning Binary Classifiers}
\label{subsubsec:classifiers}

As our feature tweaking algorithm is designed to work on tree-based ensemble classifiers, we train the following learning models to find our best estimate $\hat{f}$: Decision Trees (\dt)~\cite{Quinlan1986DT}, which can be thought of as a special case of an ensemble with a single tree; Gradient Boosted Decision Trees (\gbdt)~\cite{friedman02stochastic}, and Random Forests (\rf)~\cite{Breiman2001RF}.

The original dataset $\mathcal{D} = \{(\mathbf{x_1}, y_1), (\mathbf{x_2}, y_2), \ldots, (\mathbf{x_m}, y_m)\}$ is split into two datasets $\mathcal{D}_{\text{train}}$ and $\mathcal{D}_{\text{test}}$ using stratified random sampling.
$\mathcal{D}_{\text{train}}$ is used for training the models and accounts for 80\% of the total number of instances in $\mathcal{D}$, whilst $\mathcal{D}_{\text{test}}$ contains the remaining held-out portion used for evaluating the models.
$\mathcal{D}_{\text{train}}$ is also used to perform model selection, which is achieved by tuning the hyperparameters specific for each.

With every combination of model and corresponding hyperparameters, we run a 10-fold cross validation to find the best settings for each model -- \ie, the one with the best cross validation performance.
We measure this performance using the \emph{Area Under the Curve of the Receiver Operating Characteristic} (ROC AUC).
Each model is in turn re-trained on the whole $\mathcal{D}_{\text{train}}$ using the best hyperparameter setting.
Finally, the overall best model is deemed to be the one that best performs on the test set $\mathcal{D}_{\text{test}}$.

\subsubsection{Labelled Dataset of Ads}
\label{subsubsec:dataset}

We collect a random sample of 1,500 ads served by the \emph{Yahoo Gemini} ad network on a mobile app during one month.
To ensure reliable estimates of dwell time, we only consider ads clicked at least 500 times.
We saw that the distribution was skewed with around 80\% of the instances having an average dwell time within approximately 100 seconds, whereas the remaining 20\% sat in the long tail of very long dwell times. 
The median $\tau$ of those averages, calculated as described
in Section~\ref{subsec:adquality-definition}, is equal to $\approx 62.5$ seconds.
We therefore reach an evenly balanced ground truth, where 50\% of the instances have an average dwell time at most equal to $\tau$ and the remaining 50\% above $\tau$.

To build our labelled dataset $\mathcal{D}$, for each ad we extract the features listed in Section~\ref{subsubsec:features}.
As these are a mix of categorical (\ie, discrete) and continuous features, we apply \emph{one-hot encoding} to transform each $k$-valued categorical feature into a $k$-dimensional binary vector.
The $i$-th component of such a vector evaluates to 1 if and only if the value of the original feature is $i$, and 0 otherwise.
We also standardise continuous features by transforming their original values into their corresponding \emph{z-scores}~\cite{Kreyszig1979}.
Finally, we obtain a set of 45 features.

\subsubsection{Offline Evaluation}
\label{subsubsec:offlineval}

From the balanced labelled dataset above we derive two random partitions $\mathcal{D}_{\text{train}}$ and $\mathcal{D}_{\text{test}}$, which contain 80\% and 20\% of the total samples respectively.
We again validate our performance by running a 10-fold cross validation on $\mathcal{D}_{\text{train}}$ selecting between different models and their varying hyperparameter settings.

For \dt, we test two different node-splitting criteria $s$: \emph{Gini index} and \emph{entropy}. We also set the maximum depth of the tree $d$ to the total number of features.
For \gbdt, we use four values of the number $K$ of base trees in the ensemble, with $K =\{10, 100, 500, 1$,$000\}$ in combination with the learning rates $\alpha$,  0.001, 0.01, 0.05, 0.1, 1.
Finally, for \rf\ we test the same number of base trees as for \gbdt\ whilst again bounding the maximum depth of each base tree to the total number of features.
For each learning model we report in Table~\ref{tab:cv-performance} the hyperparameter settings leading to the best cross validation ROC AUC. The overall best performing model is \rf\ with an ensemble of $1$,$000$ base trees and maximum depth 16.

\begin{table}[t]
\centering
\small
\begin{tabular}[c]{|c||c|c|c|c|c|}
  \hline
Model & $s$ & $\alpha$ & $d$ & $K$ & ROC AUC 
\\ \hline
\rf & $-$ & $-$ & $16$ & $1$,$000$ & {\bf 0.93}
\\ \hline
\gbdt & $-$ & $0.1$ & $-$ & $100$ & 0.92
\\ \hline
\dt & \emph{entropy} & $-$ & $3$ & $-$ & 0.84
\\ \hline
 \end{tabular}
\caption{\label{tab:cv-performance} {\small Best cross validation hyperparameter settings for each learning model (``$-$'' if the hyperparameter is not considered).}}
\end{table}

To avoid mixing model selection with model evaluation, we re-train each model on $\mathcal{D}_{\text{train}}$ using its best hyperparameter setting, and assess its validity on the held-out and unseen test set $\mathcal{D}_{\text{test}}$.
We measure two standard quality metrics, $\text{F}_{\text{1}}$ and Matthews Correlation Coefficient (MCC)~\cite{matthews11comparison}.
Table~\ref{tab:test-performance} shows the results. 
\rf\ is the best performing model also with respect to the ability of generalising its predictive power to previously unseen examples.

\begin{table}[t]
\centering
\small
\begin{tabular}[c]{c|c|c|c|}
  \cline{2-4}
& \rf & \gbdt & \dt
\\ \hline
\multicolumn{1}{|c|}{$\text{F}_{\text{1}}$} & {\bf 0.84} & $0.81$ & $0.75$
\\ \hline
\multicolumn{1}{|c|}{MCC} & {\bf 0.66} & $0.63$ & $0.49$
\\ \hline
  \end{tabular}
\caption{\label{tab:test-performance} {\small Evaluation of best performing models on the test set $\mathcal{D}_{\text{test}}$.}}
\end{table}

Compared with the results reported in~\cite{Lalmas:2015:PPP:2783258.2788581}, we notice a remarkable increase of ROC AUC (+10.7\%) and a small improvement of $\text{F}_{\text{1}}$ (+1.2\%). 
In their work, Logistic Regression (\lr)~\cite{yu11dual} was the best model (ROC AUC = 0.84 and $\text{F}_{\text{1}}$ = 0.83). Our increased performance comes from a combination of a more rigorous procedure for determining the threshold $\tau$,\footnote{\small{In their work, the threshold was set as the \emph{flat} median of the observed dwell times of all ads.}} a more effective learning model, \rf, and a larger set of features.

\begin{figure}[t]
	\centering
	\includegraphics[width=\columnwidth]{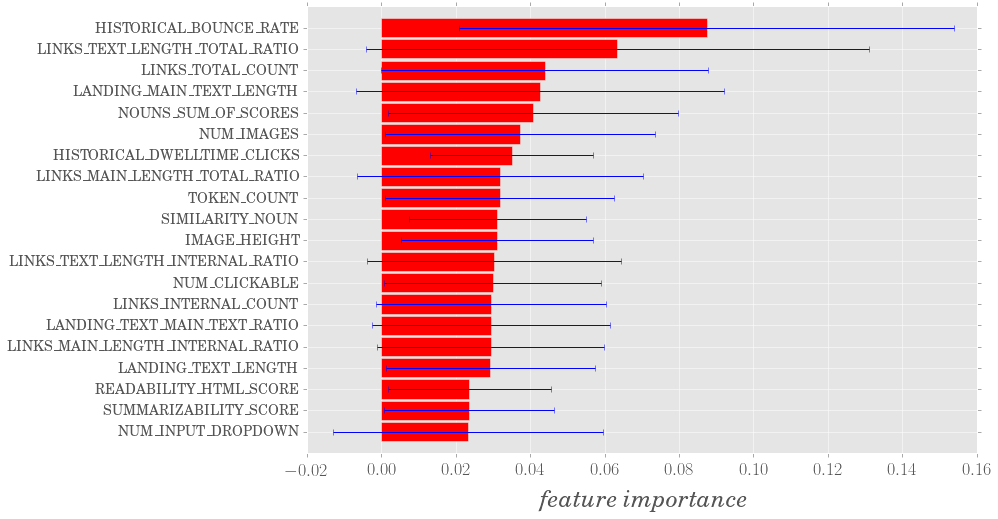}
	\caption{{\small Top-20 most important features of our \rf\ model.\label{fig:featimp}}}
\end{figure}

We also calculate the ``importance'' of each feature from the learned \rf\ model. Figure~\ref{fig:featimp} lists the top-20 most important ones.
As in~\cite{DBLP:conf/www/BarbieriSL16,Lalmas:2015:PPP:2783258.2788581}, historical features have significant predictive power. 
We keep historical features because we want the learned model for which we run our feature tweaking algorithm to be the most effective possible.
However, our feature tweaking algorithm will ignore historical features when generating recommendations, as it only considers \emph{adjustable} ad features, \ie, features that advertisers can actively alter to improve the quality of their ads.

From now on, we use \rf\ as our learned model to generate actionable suggestions on which features to tweak to turn a low quality ad into a high quality one.


\section{Experiments: Ad Feature Recommendations}
\label{sec:ad-recommendations}
\begin{figure*}[!t]
		\centering
		\subfloat[]{\includegraphics[width=0.5\columnwidth]{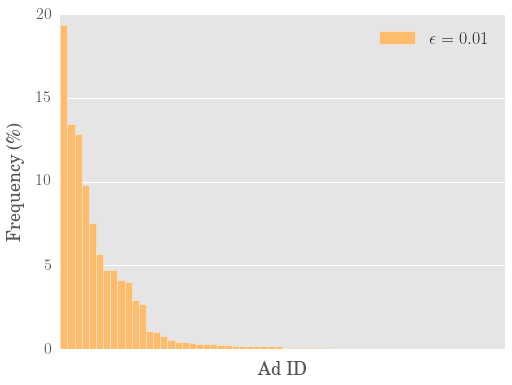}\label{fig:transf_per_ad_distr_001}}
		~
		\subfloat[]{\includegraphics[width=0.5\columnwidth]{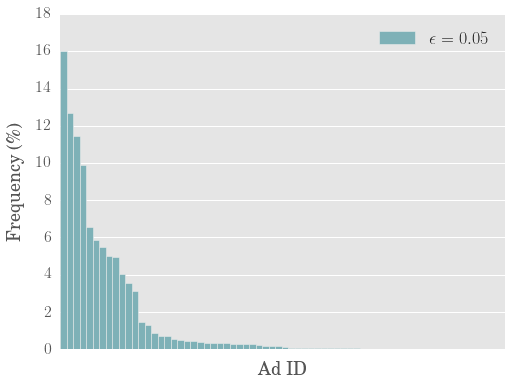}\label{fig:transf_per_ad_distr_005}}
		~
		\subfloat[]{\includegraphics[width=0.5\columnwidth]{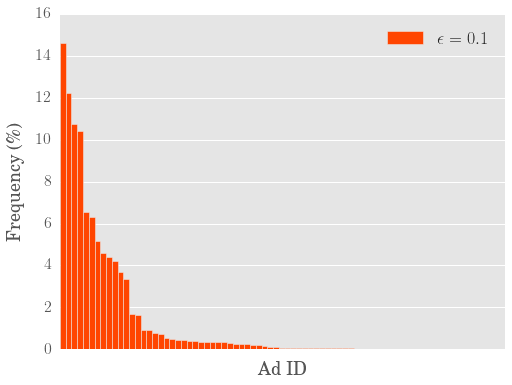}\label{fig:transf_per_ad_distr_01}}
		~
		\subfloat[]{\includegraphics[width=0.5\columnwidth]{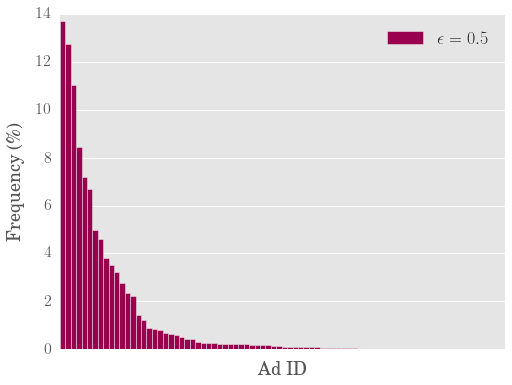}\label{fig:transf_per_ad_distr_05}}
		\caption{{\small Distribution of per-ad $\epsilon$-transformations.}\label{fig:transf_distr}}
\end{figure*}

We validate the recommendations generated with our approach, by applying our feature tweaking algorithm to our learned \rf\ model. 
Any $\mathbf{x}'$ that results from a valid (\ie, positive) $\epsilon$-trans\-for\-ma\-tion of the original negative instance $\mathbf{x}$ encapsulates a set of directives on how to positively change the ad features. 
We compute the vector $\mathbf{r}$ resulting from the component-wise difference between $\mathbf{x}'$ and $\mathbf{x}$, which is $\mathbf{r}[i] = \mathbf{x}'[i] - \mathbf{x}[i]$.
Then for each feature $i$, such that $\mathbf{r}[i] \neq 0$ (\ie, $\mathbf{x}'[i] \neq \mathbf{x}[i]$), this vector provides the \emph{magnitude} and the \emph{direction} of the changes that should be made on feature~$i$. 
The magnitude denotes the absolute value of the change (\ie, $|\mathbf{x}'[i] - \mathbf{x}[i]|$), whilst the direction indicates whether this is an \emph{increase} or a \emph{decrease} of the original value of feature $i$ (\ie, $\text{sgn}(\mathbf{x}'[i] - \mathbf{x}[i])$). 
Finally, to derive the final list of recommendations, we sort $\mathbf{r}$ according to the feature ranking, as shown in Figure~\ref{fig:featimp}.

\subsection{The impact of hyperparameters $\delta$ and $\epsilon$}
\label{subsec:rec-hyperparams}
Our approach depends on a \emph{tweaking cost} ($\delta$) associated with transforming a negative instance (low quality ad) into a positive instance (high quality ad), and a  \emph{tweaking tolerance} ($\epsilon$) used to change each individual ad feature.
We first explore how $\epsilon$ impacts on the ad \emph{coverage}, which is the percentage of ads for which our approach is able to provide recommendations.
We experiment with five values of $\epsilon$: 0.01, 0.05, 0.1, 0.5, and 1. These values can be thought of as multiples of a unit of standard deviation from each individual feature mean, as discussed in Section~\ref{subsec:tweaking}.
Table~\ref{tab:coverage} shows the highest coverage is when $\epsilon = 0.5$.

\begin{table}[t]
\centering
\small
\begin{tabular}[c]{|c|c|c|c|c|c|}
  \hline
$\epsilon$ & 0.01 & 0.05 & 0.10 & 0.50 & 1.00
\\ \hline
\emph{ad coverage} (\%) & 58.5 & 64.2 & 72.3 & {\bf 77.4} & 63.2
\\ \hline
 \end{tabular}
\caption{\label{tab:coverage} {\small The impact of tolerance threshold $\epsilon$ on ad coverage.}}
\end{table}

\begin{figure*}[ht!]
		\centering
		\subfloat[]{\includegraphics[width=0.9\columnwidth]{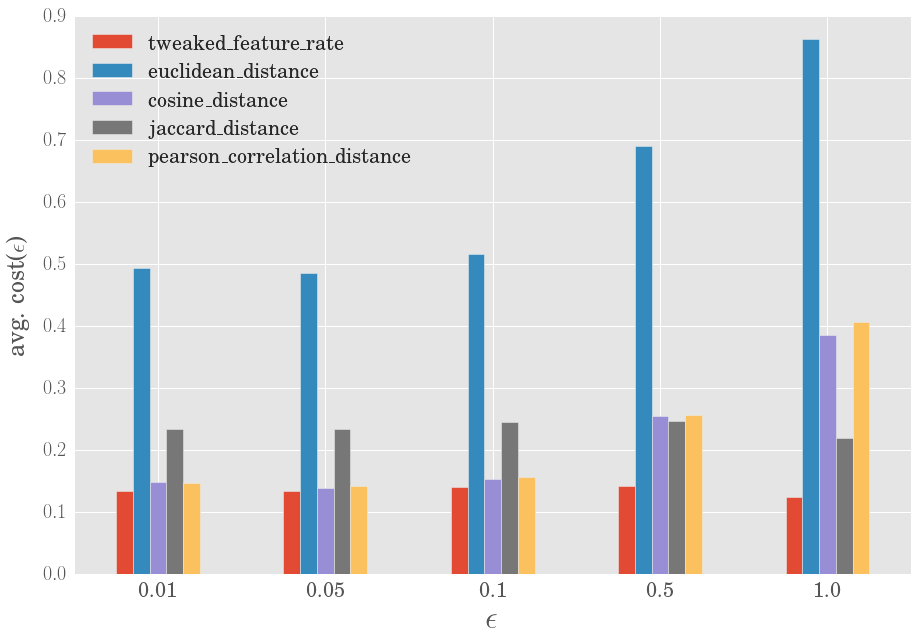}\label{fig:avg_costs_per_epsilon}}
		~
		\subfloat[]{\includegraphics[width=0.9\columnwidth]{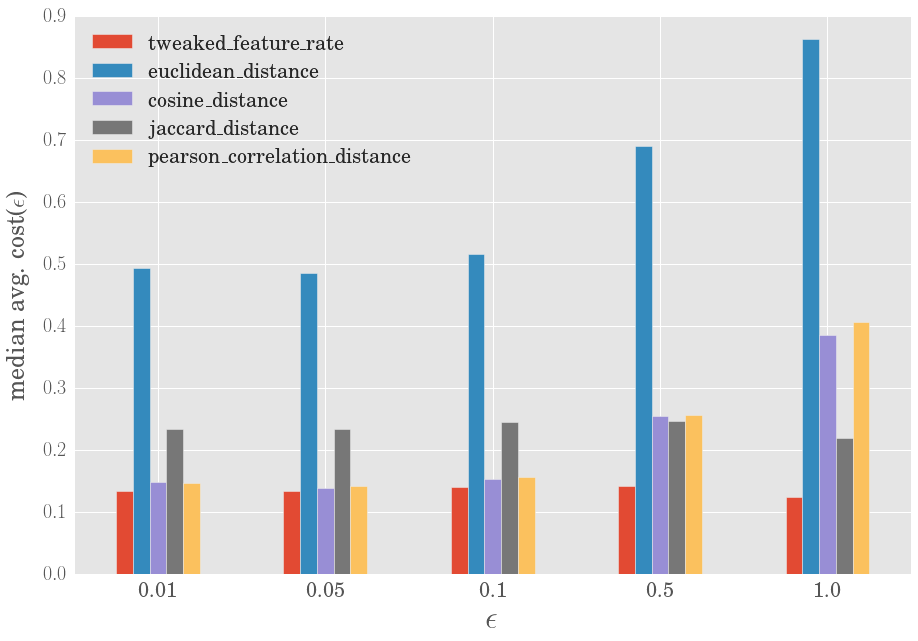}\label{fig:median_avg_costs_per_epsilon}}
		\caption{{\small The impact of tolerance threshold $\epsilon$ on costs $\delta$.}\label{fig:costs_per_epsilon}}

\end{figure*}

Although some low quality ads cannot be transformed, those that can are often associated with multiple transformations.
Figure~\ref{fig:transf_distr} shows the distribution of $\epsilon$-transformations across the set of ads, generated using different values of~$\epsilon$ (except $\epsilon=1$ which is similar to $\epsilon=0.05$).
All the distributions are skewed offering a high number of transformations proposed for few ads. 
Interestingly, the number of transformations is more evenly distributed across the ads when $\epsilon$ increases.
This is in agreement with the finding above, where larger values of $\epsilon$ result in a higher coverage before decreasing again between $\epsilon=0.5$ and 1.

To choose the most appropriate transformation for an ad, we experiment with several tweaking cost functions $\delta$, each taking as input the original ($\mathbf{x}$) and the transformed ($\mathbf{x}'$) feature vectors:
\begin{itemize}\itemsep0pt
	\item {\sf tweaked\_feature\_rate}: proportion of features affected by the transformation of $\mathbf{x}$ into $\mathbf{x}'$ (range = [0, 1]);
	\item {\sf euclidean\_distance}: Euclidean distance between $\mathbf{x}$ and $\mathbf{x}'$ (range = $\R$);
	\item {\sf cosine\_distance}: 1 minus the cosine of the angle between $\mathbf{x}$ and $\mathbf{x}'$ (range = [0, 2]);
	\item {\sf jaccard\_distance}: one's complement of the Jaccard similarity between $\mathbf{x}$ and $\mathbf{x}'$ (range = [0, 1]);
	\item {\sf pearson\_correlation\_distance}: 1 minus the Pearson's correlation coefficient between $\mathbf{x}$ and $\mathbf{x}'$ (range = [0, 2]).
\end{itemize}

Up to a certain value, the tolerance $\epsilon$ is positively correlated with the ad coverage. 
We explore how it impacts the five tweaking cost functions.
Figure~\ref{fig:costs_per_epsilon}(a) plots the micro-average costs and Figure~\ref{fig:costs_per_epsilon}(b) shows the median of all the individual per-ad average costs.
In general, the greater the tolerance the higher the cost (except for {\sf tweaked\_feature\_rate} and {\sf jaccard\_distance} when $\epsilon=1$); thus a trade-off between $\epsilon$ (\ie, ad coverage) and the cost of ad transformations $\delta$ is desirable.

\subsection{Evaluating Recommendations}
\label{subsec:rec-eval}

\begin{figure*}[htb!]
	\subfloat[]{\label{fig:top-1}\includegraphics[width=0.33\textwidth]{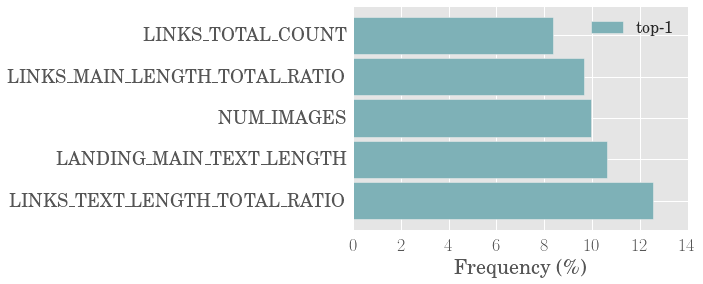}}
	~
	\subfloat[]{\label{fig:top-2}\includegraphics[width=0.33\textwidth]{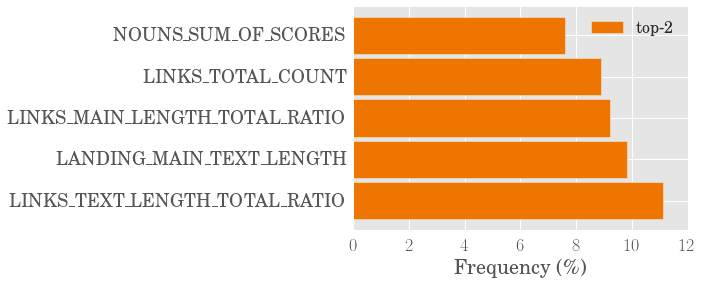}}
	~
	\subfloat[]{\label{fig:top-3}\includegraphics[width=0.33\textwidth]{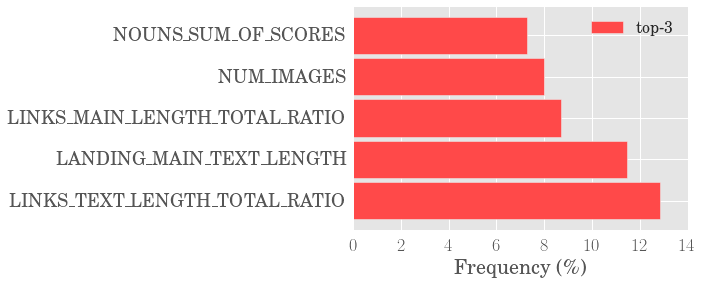}}
\caption{{\small Top-5 most frequent features appearing in the top-1, top-2 and top-3 $\epsilon$-transformations ($\epsilon=0.05$).}\label{fig:top-k}}
\end{figure*}

We first present descriptive statistics on the recommendations obtained with our approach on a set of 100 low quality ad landing pages, the true negative instances in  $\mathcal{D}_{\text{test}}$.
Each recommendation either suggests to \emph{increase} or \emph{decrease} the value of a given feature.
Overall, the recommendations are almost evenly distributed over the two cases above.

In Figure~\ref{fig:top-k}, we list the top-5 most frequent features recommended to be tweaked according to the top-1, top-2 and top-3 proposed $\epsilon$-transformations by measuring the relative frequency of each feature appearing among each of the $\epsilon$-transformations. The most frequent feature is {\small LINKS\-\_TEXT\-\_LENGTH\-\_TOTAL\-\_RATIO} in all settings, whi\-ch measures the ratio of text length to the total number of hyperlinks in the ad landing page.
Interestingly, \emph{all} the recommendations concerning this feature suggest to \emph{decrease} its value.
This indicates that low quality ad landing pages generally exhibit an unbalanced ratio of text to hyperlinks suggesting that saturating a page in links rather than content has negative effects on dwell time.

We measure the Pearson's correlation coefficient ($\rho$) between feature rankings appearing in the top-1, top-2 and top-3 $\epsilon$-trans\-for\-ma\-tions.
All three rankings are strongly related with each other, with top-1 reaching $\rho$ = 0.93 and 0.81 when compared to top-2 and top-3, respectively. Similarly, top-2 is highly correlated to top-3 ($\rho$ = 0.79).\footnote{\small{All values are statistically significant at $\alpha$ = 0.01}.}
We also compute the correlation coefficient between top-1 $\epsilon$-transformations for all values of $\epsilon$, and $\delta=\text{{\sf cosine\_distance}}$.
The top-1 rankings derived from $\epsilon$ = 0.05 and 0.1 are the highest correlated ($\rho$ = 0.92).
However, there is no statistical significant correlation between top-1 rankings when $\epsilon$ = 0.1 and 0.5, indicating that higher values of tolerance may impact more on the features requiring change.

We now perform a \emph{qualitative} and \emph{quantitative} assessment of such recommendations. 
For each low quality landing page, we focus on its top-$k$ $\epsilon$-transformations, \ie, the $k$ less costly transformations according to the cost function $\delta$.
In turn, each transformation contains a list of recommendations, sorted by the feature rank they refer to.
We set the hyperparameters to $\epsilon=0.05$ and $\delta=$ {\sf cosine\_distance}, as this combination provides the best trade-off between ad coverage and average cost. We consider the top-3 $\epsilon$-transformations suggested for each ad landing page.
Around $91.0\%$ of landing pages can be associated with all three $\epsilon$-transformations, whereas our algorithm provides the remaining $7.5\%$ and $1.5\%$ with two and one $\epsilon$-trans\-for\-ma\-tion, respectively.

We also asked an internal team of creative strategists (CS)\footnote{\small{Creative strategists work with advertisers' web masters on strategic choices to help them developing effective advertising messages.}} to validate the recommendations generated by our approach.
Each CS was assigned a set of ad landing pages with the corresponding $\epsilon$-trans\-for\-ma\-tions, and additional metadata useful for assessing the recommendations within each transformation.
The same set of ad landing pages -- and therefore the same list of recommendations -- was assessed by two CSs, who were asked to rate each recommendation as \emph{helpful}, \emph{non-helpful}, or \emph{non-actionable}.
A recommendation is deemed \emph{helpful} when it is likely to help the advertiser to improve the user experience of the ad, and \emph{non-helpful} otherwise.
A \emph{non-actionable} recommendation is one that cannot be practically implemented.
Whenever a disagreement occurred, a third CS was called to resolve the conflict.

Overall, $57.3\%$ of all the generated recommendations are rated helpful with an inter-agreement rate of $60.4\%$ and only $0.4\%$ result in a non-actionable suggestion.
We also look at the $42.3\%$ non-helpful recommendations, and saw that about $25\%$ can be considered ``neutral''; that is, they would not hurt the user experience if discarded as well as not adding any positive value if implemented.

Non-helpful tweaks might occur due to two reasons. First, the learned model we leverage for generating feature recommendations -- no matter how accurate it is -- is not perfect; therefore, a true negative instance that is transformed into a positive prediction does not necessarily mean it is \emph{actually} positive.
Second, tuning the hyperparameters ($\delta$ and $\epsilon$) of our algorithm affects the set of candidate transformations. 
As such, limiting non-helpful tweaks can be achieved by improving the accuracy of the learned model and choosing values of the hyperparameters to minimize errors.

Furthermore, when we further look into the non-actionable recommendations we see that these are related to the features {\small ADULT\_SCORE} and {\small NUM\_INPUT \_DROPDOWN}. 
Our algorithm suggests to decrease the value of those features; however, the ad landing pages do not contain adult words nor drop-downs.
Most likely, the ad copies and landing pages used to generate recommendations have changed before the CSs performed their assessment. 

\begin{figure}[t]
	\centering
	\includegraphics[width=\columnwidth]{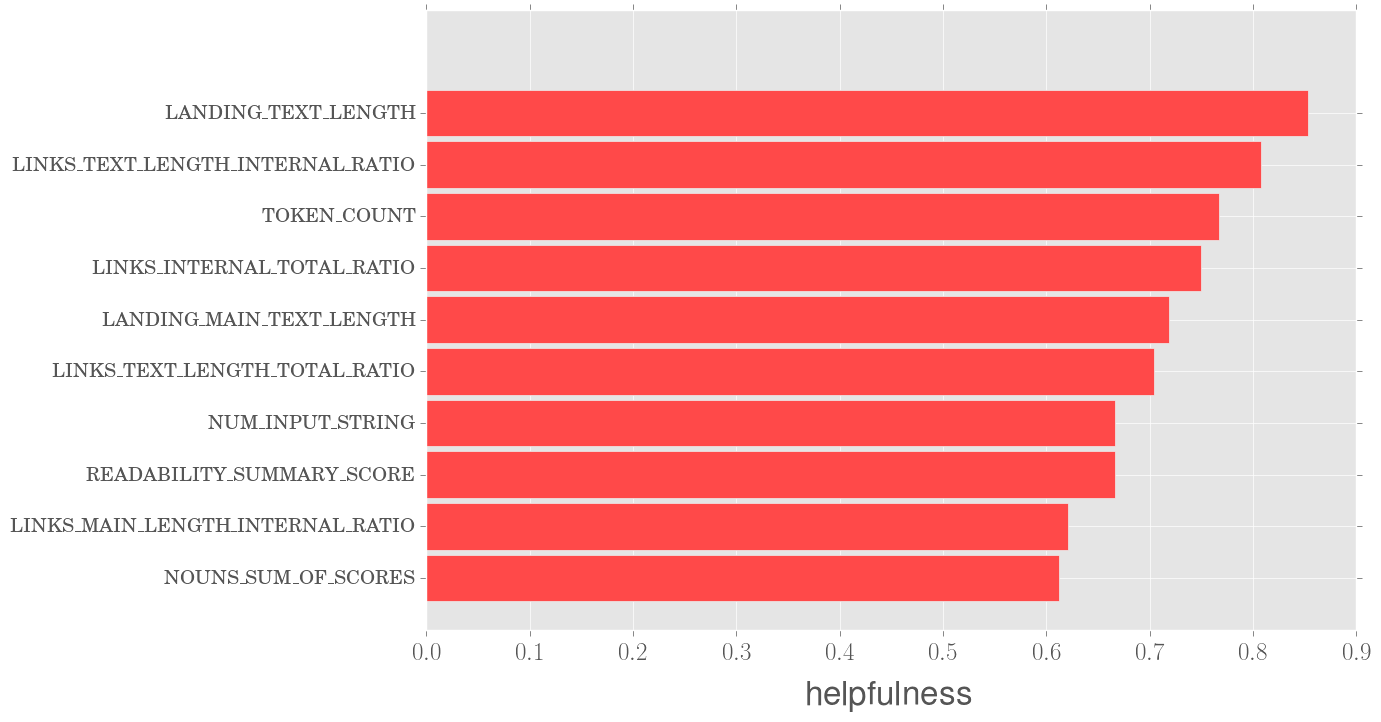}
\caption{{\small Top-10 most helpful feature recommendations according to the {\sf helpfulness} score.}
\label{fig:helpfulness}}
\end{figure}

Finally, we measure the ``helpfulness'' of each feature recommendation as follows:
\[
\text{{\sf helpfulness}}(i) = \frac{|\text{{\sf helpful}}(i)|}{|\text{{\sf helpful}}(i)|+|\lnot \text{{\sf helpful}}(i)|}
\]
This computes the relative frequency of recommendations for feature $i$ as being described as helpful by the CS team.
In Figure~\ref{fig:helpfulness}, we report the ranked list of features involved in the top-10 most helpfulness recommendations.
A similar ranking is obtained if we weight the helpfulness score on the basis of the \emph{overall} relative recommendation frequency.
The majority of the most helpful recommendations were features extracted from the DOM structure and content of the ad landing page, indicating that high quality landing pages should exhibit a good balance between textual content and hyperlinks.
Those features were the most predictive in our \rf\ ad quality model (Figure~\ref{fig:featimp}).

\section{Related Work}
\label{sec:related}

The research challenge addressed in this work is largely unexplored. Although machine learning has received a lot of attention in recent years, the focus has been mainly on the accuracy, efficiency, scalability, and robustness of the proposed various techniques.
Works on extracting actionable knowledge from machine-learned models have been mostly
conducted within the business and marketing domains. 
Early works have focused on the development of interestingness metrics as proxy measures of knowledge actionability~\cite{hilderman2000pkdd, cao2007ijbidm}.

Another line of research on actionable knowledge discovery concerns post-processing techniques. Liu \etal\ propose methods for pruning and summarizing learned rules, as well as matching rules by similarity~\cite{liu1996aaai, liu1999kdd}.
Cao \etal\ present domain-driven data
mining; a paradigm shift from a research-centered discipline to a practical tool for actionable knowledge~\cite{cao2006pakdd, cao2010tkde}.
The authors discuss several frameworks for handling different problems and applications.

Many works discuss post-processing techniques specifically tailored to decision trees~\cite{yang2003icdm, karim2013jsea, yang2007tkde, du2011lori}. Yang \etal\ study the problem of proposing actions to maximise the expected profit for a group of input instances based on a single decision tree, and introduce a greedy algorithm to approximately solve such a problem~\cite{yang2003icdm}.
This is significantly different
from our work; in fact, our work is more related to the one presented by Cui \etal~\cite{cui2015kdd}.
Here, the authors propose a method to support actionability for additive tree models (ATMs), which is to find the set of actions that can change the prediction of an input instance to a desired status with the minimum cost. The authors formulate the problem as an instance of integer linear programming (ILP) and solve it using existing techniques.

Similarly to Cui \etal, we also consider transforming the prediction for a given instance output by an ensemble of trees, and we introduce an algorithm that finds the \emph{exact} solution to the problem.
Our work differs from theirs in several aspects:
\emph{(i)} We tackle the theoretical intractability (NP-hardness) of the problem by designing an algorithm that creates a feedback loop with the original model to build a set of candidate transformations without the need, in practice, to explore the entire exponential search space;
\emph{(ii)} We introduce another hyperparameter ($\epsilon$) to  govern the amount of change that each feature can sustain;
\emph{(iii)} We experiment with five concrete functions describing the cost of each transformation ($\delta$);
\emph{(iv)} We leverage on the importance of each feature derived from the model to rank the final list of recommendations;
\emph{(v)} We focus on the actual recommendations generated, and how they impact in practice on a real use case, if properly implemented.

More recent work on related topics are those of Ribeiro \etal~\cite{Ribeiro:2016:WIT:2939672.2939778, ribeiro2016model}. In particular, \cite{Ribeiro:2016:WIT:2939672.2939778} presents LIME, a method that aims to explain the predictions of any classifier by learning an interpretable model that is specifically built around the predictions of interest.
They frame this task as a submodular optimization problem, which the authors solved using a well-known greedy algorithm achieving performance guarantees.
They test their algorithm on different models for text (\eg, random forests) and image classification (\eg, neural networks), and validate the utility of generated explanations both via simulated and human-assessed experiments.


\section{Conclusions}
\label{sec:conclusions}

Machine-learned models are often designed to favour accuracy of prediction at the  expense of human-interpretability.
However, in some circumstances it becomes important to understand why the model returns a certain prediction on a given instance and how such an instance could be transformed in such a way that the model changes its original prediction. We investigate this problem within the context of general ensembles of tree-based classifiers, which has been proven to be NP-hard. We then introduce an algorithm that is able to transform a true negative instance into a set of new ``proposed'' positive instances by shifting their position in the feature space. 
The algorithm leverages the internals of the learned ensemble to \emph{tweak} the feature-based representation of a true negative instance so that the new ``proposed'' ones are promoted to a positive classification when re-input to the classifier. 

Despite computationally intractable in the worst case, we demonstrate the applicability of our approach on a real-world use case in online advertising.  The feasibility of our approach has been achieved by \emph{(i)} setting an upper bound to the maximum number of changes affecting each instance (\ie, at most equivalent to the number of features), which can be controlled at training time, and by \emph{(ii)} creating a feedback loop with the original model to build a set of candidate transformations without the need, in practice, to explore the entire exponential search space.  

After designing an effective Random Forest classifier able to separate between \emph{low} and \emph{high} quality ads -- our application scenario -- we automatically provide ``actionable'' suggestions on how to optimally convert a low quality ad (negative instance) into a high quality one (positive instance) using our approach. To illustrate the outcomes of our algorithm, we assess the quality of the recommendations that our method generates from a dataset of ads served by a large ad network, \emph{Yahoo Gemini}.
An evaluation conducted by an internal team of creative strategists shows that 57.3\% of the provided recommendations are indeed helpful, and likely to improve the ad quality, if implemented.

In future work, we plan to extend the approach presented in this work to multi-class setting as well as to other learning models, and to encapsulate it into a reinforcement learning framework.

\begin{acks}
The authors would like to thank Huw Evans, Mahlon Chute, and all the Yahoo's internal team of creative strategists for their invaluable contributions in evaluating the quality of the method presented in this work.
\end{acks}




\end{document}